\documentclass[9pt,twocolumn,twoside]{pnas-new}

\usepackage{tikz}
\usepackage{verbatim}
\usepackage{tabularx}
\usepackage{xr}
\usetikzlibrary{fit, backgrounds, matrix, arrows.meta}
\makeatletter
\newcommand*{\addFileDependency}[1]{
  \typeout{(#1)}
  \@addtofilelist{#1}
  \IfFileExists{#1}{}{\typeout{No file #1.}}
}
\usepackage{algorithm} 
\usepackage{algpseudocode} 
\algnewcommand\algorithmicforeach{\textbf{for each}}
\algdef{S}[FOR]{ForEach}[1]{\algorithmicforeach\ #1\ \algorithmicdo}
\makeatother

\newcommand*{\myexternaldocument}[1]{%
    \externaldocument{#1}%
    \addFileDependency{#1.tex}%
    \addFileDependency{#1.aux}%
}
\myexternaldocument{SI}

\templatetype{pnasresearcharticle} 

\title{Automating Transfer Credit Assessment in Student Mobility - A Natural Language Processing-based Approach}

\author[a,1]{Dhivya Chandrasekaran}
\author[a,]{Vijay Mago} 

\affil[a]{Department of Computer Science, Lakehead University, 955 Oliver Road, Thunder Bay, P7B5E1 }

\leadauthor{Chandrasekaran} 

\significancestatement{With the increase in the number of student mobility across institutions, the focus on ensuring a seamless transition has gained significance. Since manual assessment of transfer credit is labor-intensive and biased, this article attempts to leverage the recent advancements in the field of Natural Language Processing to automate the assessment of transfer credits offered by institutions to incoming students. Taking into consideration the structure and semantics of learning outcomes, the proposed model aggregates the similarity between learning outcomes to determine course level similarity. To offer flexibility to decision-making authorities, three tunable parameters are provided to adjust leniency. The article also discusses the various challenges facing future research in applying NLP techniques in the field of transfer and articulation. 
}

\authorcontributions{D.C. conceived and designed the analysis, collected the data and wrote the paper, V.M is the principal supervisor.}
\authordeclaration{The authors declare no competing interest}
\correspondingauthor{\textsuperscript{1} To whom correspondence may be addressed. E-mail: dchandra@lakeheadu.ca}

\keywords{Natural Language Processing $|$ Semantic Similarity $|$ Articulation Agreements $|$ Higher Education} 

\begin{abstract}
Student mobility or academic mobility involves students moving between institutions during their post-secondary education, and one of the challenging tasks in this process is to assess the transfer credits to be offered to the incoming student. In general, this process involves domain experts comparing the learning outcomes of the courses, to decide on offering transfer credits to the incoming students. This manual implementation is not only labor-intensive but also influenced by undue bias and administrative complexity. The proposed research article focuses on identifying a model that exploits the advancements in the field of Natural Language Processing (NLP) to effectively automate this process. Given the unique structure, domain specificity, and complexity of learning outcomes (LOs), a need for designing a tailor-made model arises. The proposed model uses a clustering-inspired methodology based on knowledge-based semantic similarity measures to assess the taxonomic similarity of LOs and a transformer-based semantic similarity model to assess the semantic similarity of the LOs. The similarity between LOs is further aggregated to form course to course similarity. Due to the lack of quality benchmark datasets, a new benchmark dataset containing seven course-to-course similarity measures is proposed. Understanding the inherent need for flexibility in the decision-making process the aggregation part of the model offers tunable parameters to accommodate different scenarios. While providing an efficient model to assess the similarity between courses with existing resources, this research work steers future research attempts to apply NLP in the field of articulation in an ideal direction by highlighting the persisting research gaps.

\end{abstract}

\dates{This manuscript was compiled on \today}
\doi{\url{www.pnas.org/cgi/doi/10.1073/pnas.XXXXXXXXXX}}

\begin{document}

\maketitle
\thispagestyle{firststyle}
\ifthenelse{\boolean{shortarticle}}{\ifthenelse{\boolean{singlecolumn}}{\abscontentformatted}{\abscontent}}{}

\dropcap{W}ith the significant growth in the enrollment of students in post-secondary institutions and the growing trend of interest in diversifying one’s scope of education, there is an increasing demand among the academic community to standardize the process of student mobility. Student mobility is defined as “any academic mobility which takes place within a student’s program of study in post-secondary education \cite{junor2008student}.” Student mobility could be both international and domestic. While there are various barriers to student mobility, offering transfer credits for students moving from one post-secondary institution to another is considered one of the most critical and labor-intensive tasks \cite{8685082}. Various rules and regulations are proposed and adopted by institutions across different levels (provincial, national, or international) to assess transfer credits, but most of these methods are time-consuming, subjective, and influenced by undue human bias. The key parameter used in assessing the similarity between programs or courses across institutions is learning outcome (LO), which provides context on the competencies; achieved by students on completion of a respective course or program. To standardize this assessment, LOs are categorized into various levels based on Bloom’s taxonomy. Bloom’s taxonomy proposed by Benjamin Bloom \cite{bloom1956taxonomy} attempts to classify the learning outcomes into six different categories based on their “complexity and specificity”, namely \textit{knowledge, comprehension, application, analysis, synthesis, and evaluation.}

Semantic similarity, being one of the most researched Natural language processing (NLP) tasks, has seen significant breakthroughs in recent years with the introduction of transformer-based language models. These language models employ attention mechanisms to capture the semantic and contextual meaning of text data and represent them as real-valued vectors, that are aligned in an embedding space such that the angle between these vectors provides the similarity between the text in consideration. In an attempt to reduce the inherent complexity and bias, and exploit the advancements in the field of NLP, we propose a model that determines the similarity between courses; based on the semantic and taxonomic similarity of their learning outcomes. The proposed model
\begin{itemize}
    \item ascertains taxonomic similarity of LOs based on Bloom’s taxonomy.
    \item determines semantic similarity of LOs using RoBERTa language model.
    \item provides a flexible aggregation method to determine the overall similarity between courses.
\end{itemize}
In the Background section, we describe the background of the student mobility process and semantic similarity techniques. In the Methodology section, where describe in detail the various components of the proposed methodology followed by the discussion of results in the Results section. In the Challenges section,  we annotate the challenges in automating the process of assessing transfer credit and conclude with the future scope of research in the final section.
\section*{Background} 
This section provides a brief overview of the student mobility process across the world and the structural organization of learning outcomes. Various semantic similarity methods developed over the years are discussed in the final sub-section thus providing an insight into the necessary concepts to understand the importance of the research and the choices made to develop the proposed methodology. 
\begin{figure*}[ht!]
    \centering
    \includegraphics[width=0.9\textwidth]{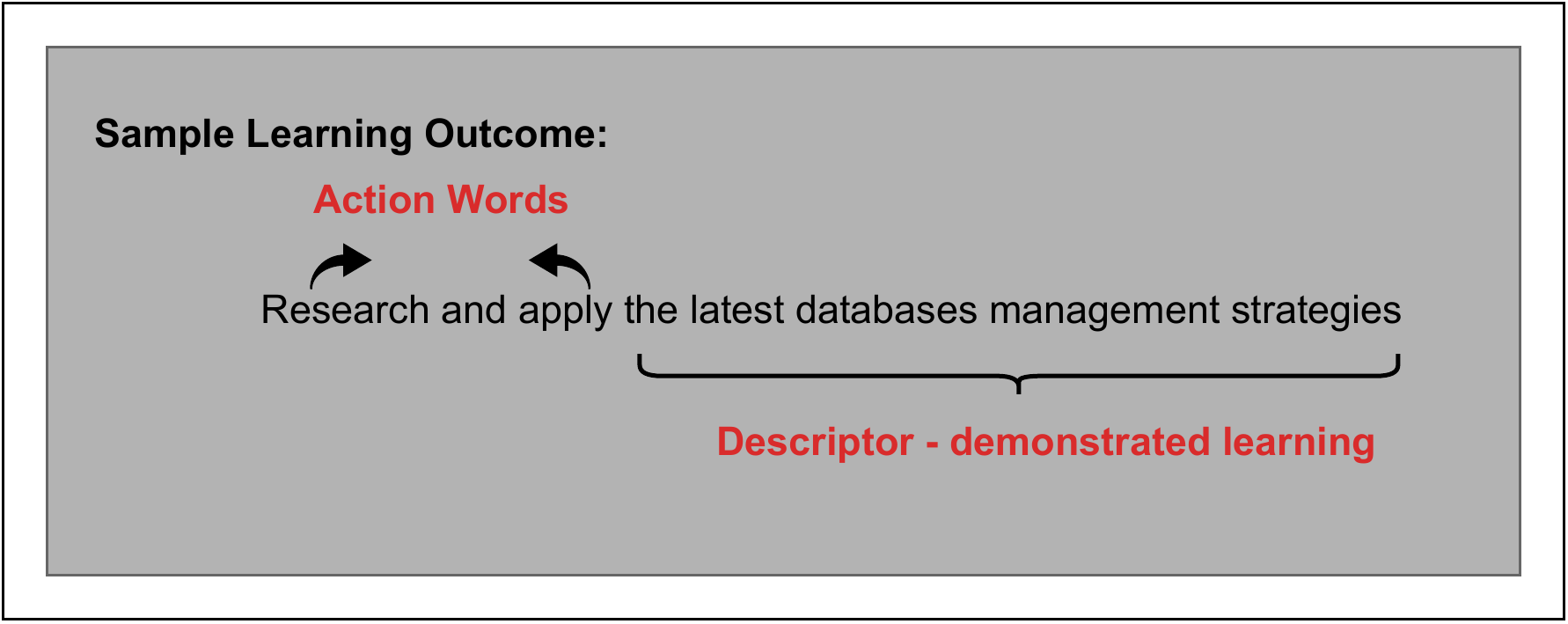}
    \caption{Learning Outcome}
    \label{fig:losample}
\end{figure*}
\subsection*{Student Mobility}
The movement of students across institutions for higher education has been in existence for decades in the form of international student exchange programs, lateral transfers, and so on. Governments across the world follow different measures to standardize the process to ensure transparency and equity for students. According to the Organisation for Economic Co-operation and Development (OCED) indicators, there are approximately 5.3 million internationally mobile students \cite{oced}. Mobile students include both international students who cross borders to pursue education and foreign students who are from a different nationality than the country of the host institution. Mobile students face a wide range of barriers both academic and non-academic. Academic barriers include the lack of necessary qualifications and non-transferability of credits, while non-academic barriers may include social, cultural, financial, and psychological barriers. Governments across the world have taken various measures to reduce these barriers to enable a smooth transition for students. The Bologna process formed as a result of the Bologna declaration of 1999, provides guidance for the European Higher Education Area comprising 48 countries in the standardization of higher education and credit evaluation. In the United Kingdom, institutions like Southern England Consortium for Credit Accumulation and Transfer (SEEC) and  Northern Universities Consortium for Credit Accumulation and Transfer (NUCCAT) oversee the collaboration between universities to allocate academic credits which are treated as currency awarded to students on completion of requirements. Canada offers provincial supervision of articulation agreements between institutions, with the provinces of British Columbia and Alberta leading and the provinces of Ontario and Saskatchewan following yet way behind. The Ontario Council on Articulation and Transfer (ONCAT) carries out funded research to explore venues to increase the agreements between universities and colleges in Ontario. The credit transfer system in the United States is decentralized and often carried out by non-profit organizations designated for this specific purpose. In Australia, the eight most prominent universities established the Go8 credit transfer agreement to offer credit to students who move between these institutions. While there are various such governances on a national level, most international credit evaluations are carried out in a need-based manner. In addition to being an academic barrier; credit evaluation also has a direct impact on one of the most important non-academic barriers - the financial barrier. Hence, all agencies offer special attention to make this process fair and accessible to the population of mobile students worldwide.

\subsection{Learning Outcomes}
\begin{figure*}
    \centering
    \includegraphics[width =1.1\textwidth]{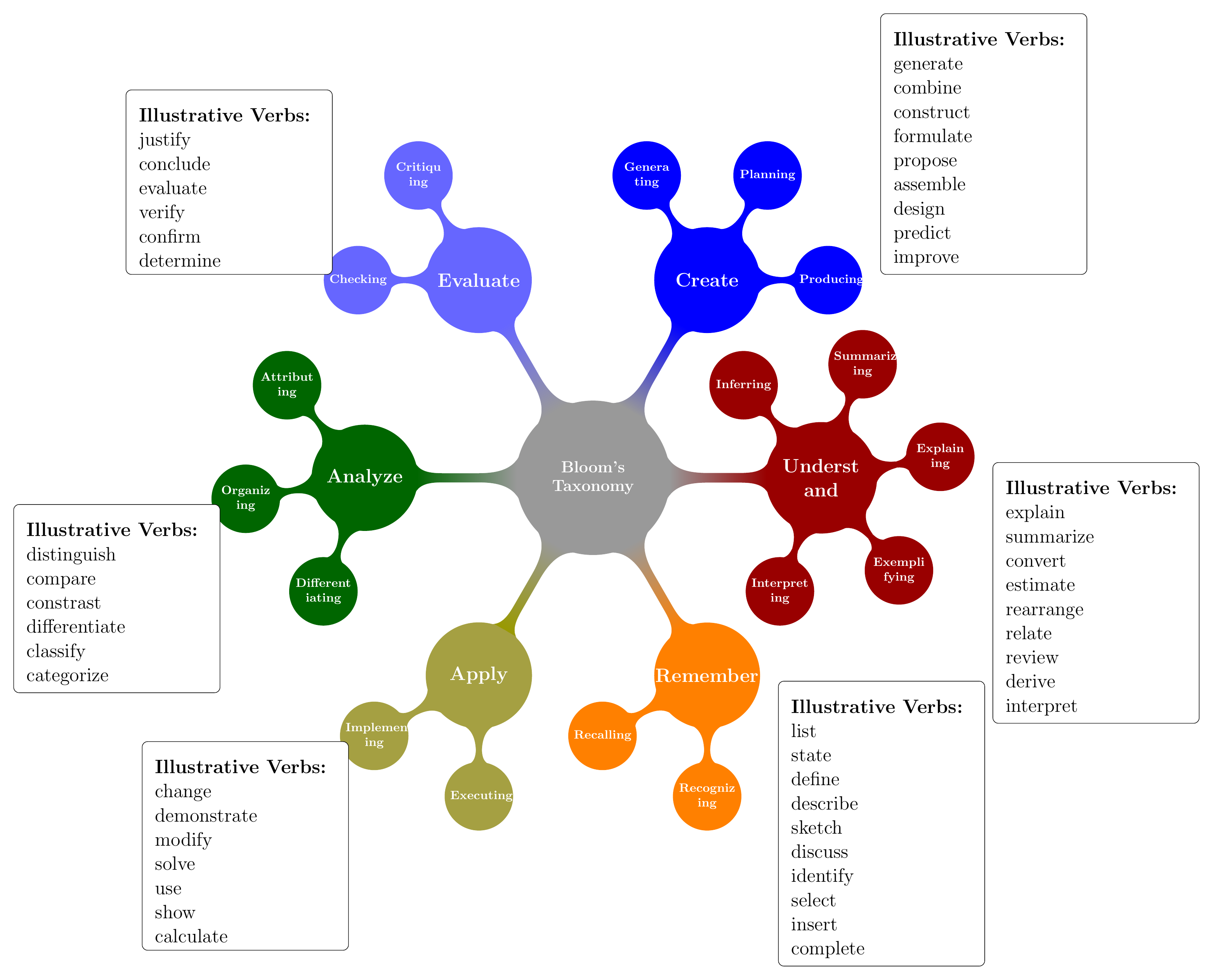}
    \caption{Bloom's taxonomy and corresponding illustrative verbs}
    \label{fig:clustering}
\end{figure*}

Credit evaluation is carried out by domain experts in the receiving institution by analyzing the learning outcomes of the courses the students have completed in their previous institution. Learning outcomes are often statements with two distinct components namely the action words and the descriptor. The descriptor part provides the knowledge the students have learned in a given course or program and the action words provide the level of competency achieved for each specific knowledge item. An example of learning outcomes for a computer programming course is provided in Fig. \ref{fig:losample}. The taxonomy for educational objectives was developed by Bloom et al. \cite{bloom1956taxonomy} and later revised by Anderson et al. \cite{anderson1994bloom}. The six levels of the original taxonomy are \textit{knowledge, comprehension, application, analysis, synthesis, and evaluation}. In the revised version, two of these levels were interchanged and three levels were renamed to provide better context to the level of the acquired knowledge. Hence the levels of the revised  taxonomy are \textit{“Remember, Understand, Apply, Analyze, Evaluate, and Create.”} Each level encompasses sub-levels of more concrete knowledge items and these are provided in Fig. \ref{fig:clustering}. In order to estimate the similarity between learning outcomes, it is important to understand their structure and organization.
\subsection{Semantic Similarity}
The semantic textual similarity (STS) is defined as the similarity in the meaning of text data in consideration. Various semantic similarity methods proposed over the past two decades can be broadly classified as knowledge-based and corpus-based methods \cite{chandrasekaran2021evolution}. Knowledge-based methods rely on ontologies like WordNet\footnote{https://wordnet.princeton.edu/}, DBPedia\footnote{https://www.dbpedia.org/}, BabelNet\footnote{https://babelnet.org/}, etc. The ontologies are conceptualized as graphs where the words represent the nodes grouped hierarchically, and the edges represent the semantic relationship between the words. Rada et al. \cite{rada1989development} followed a straightforward approach and introduced the $path$ measure in which the semantic similarity is inversely proportional to the number of edges between the two words. However, this method ignored the taxonomical information offered by the underlying ontologies. Wu et al., \cite{wu1994verbs} proposed the $wup$ measure that measured the semantic similarity in terms of the least common subsumer (LCS). Given two words, LCS is defined as the common parent they share in the taxonomy. Leacock et al. \cite{leacock1998combining} proposed the $lch$ measure by extending the $path$ measure to incorporate the depth of the taxonomy to calculate semantic similarity . The formulations of these methods are provided in Table \ref{tab:S1} in the Appendix. Other knowledge-based approaches include feature-based semantic similarity methods, that calculate similarity using the features of the words like their dictionary definition, grammatical position, etc., and information content-based methods that measure semantic similarity using the level of information conveyed by the words when appearing in a context. Corpus-based semantic similarity methods construct numerical representations of data called embeddings using large text corpora. Traditional methods like Bag of Words (BoW), Term Frequency - Inverse Document Frequency (TF-IDF) used one-hot encoding techniques or word counts to generate embeddings. These methods ignored the polysemy of text data and suffered due to data sparsity. Mikolov et al. \cite{mikolov2013efficient} used a simple neural network with one hidden layer to generate word-embeddings when used in simple mathematical formulations, produced results that were closely related to human understanding. Pennington et al. \cite{pennington2014glove} used word co-occurrence matrices and dimension reduction techniques like PCA to generate embeddings. The introduction of transformer-based language models, which produced state-of-the-art results in a wide range of NLP tasks, resulted in a breakthrough in semantic similarity analysis as well. Vaswani et al. \cite{vaswani2017attention} proposed the transformer architecture, which used stacks of encoders and decoders with multiple attention heads for machine translation tasks. Delvin et al. \cite{devlin2019bert} used this architecture to introduce the first transformer-based language model, the Bidirectional Encoder Representations from Transformers (BERT) that generated contextualized word embeddings. BERT models were pre-trained on large text corpora and further fine-tuned to a specific task. Various versions of BERT were subsequently released namely, RoBERTa \cite{liu2019roberta} - trained on a larger corpus over longer periods of time, ALBERT \cite{lan2019albert} - a lite version achieved using parameter reduction techniques, BioBERT \cite{lee2020biobert} - trained on a corpus of biomedical text, SciBERT \cite{beltagy2019scibert} - trained on a corpus of scientific articles, and TweetBERT \cite{qudar2020tweetbert} - trained on a corpus of tweets. Other transformer-based language models like T5 \cite{raffel2019exploring}, GPT \cite{radford2018improving}, GPT-2 \cite{radford2019language}, etc., use the same transformer architecture with significantly larger corpora and an increased number of parameters. Though these models achieve state-of-the-art results the computational requirements render them challenging to implement in real-time tasks \cite{strubell2019energy}.
\begin{figure*}
    \centering
    \includegraphics[width=0.7\textwidth]{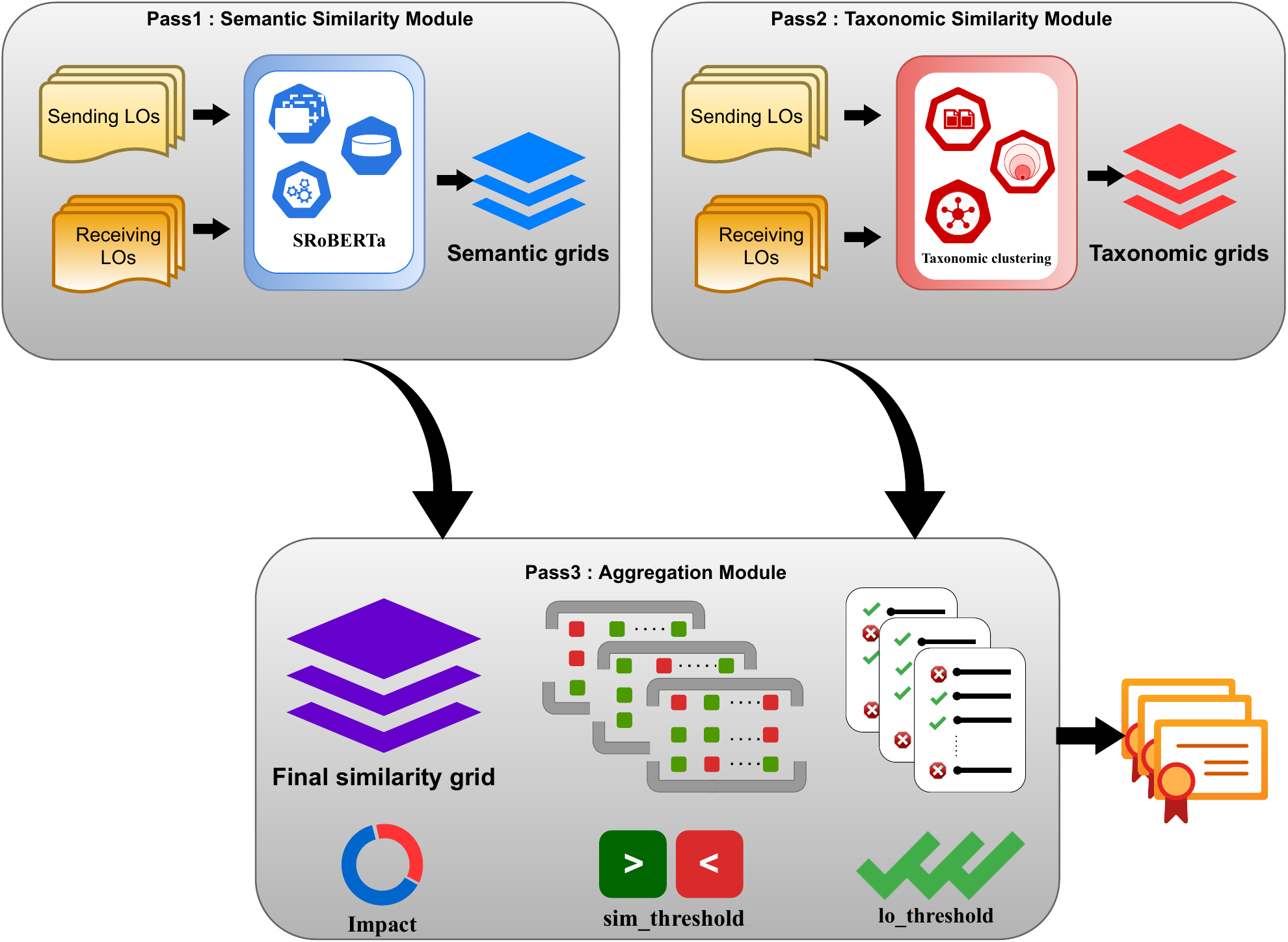}
    \caption{Proposed model architecture}
    \label{fig:archi}
\end{figure*}
\section{Methodology}
This section describes in detail the three modules of the proposed methodology namely,
\begin{itemize}
    \item Pass 1: Taxonomic similarity
    \item Pass 2: Semantic similarity
    \item Pass 3: Aggregation
\end{itemize}

Given the learning outcomes of the courses in comparison, Pass 1 generates a \textit{taxonomic\_similarity\_grid}, and Pass 2 generates a \textit{semantic\_similarity\_grid} where the rows and columns are populated with the learning outcomes, and the cells are populated with the taxonomic similarity value and the semantic similarity value. These two grids are further passed on to Pass 3 where the final similarity between learning outcomes is assessed factoring in both the similarity values and further aggregated to arrive at the course level similarity. The architecture of the proposed model is presented in Fig. \ref{fig:archi}. The pseudocode for all the three modules in the proposed model are provided in Appendix.
\subsection{Pass 1: Taxonomic similarity}
A clustering-inspired methodology is proposed to determine the taxonomic similarity between learning outcomes. Six different clusters, one for each level in Bloom’s taxonomy are initialized with a list of illustrative verbs that best describe the cognitive competency achieved, specifically in the field of engineering \cite{swart2009evaluation}, as shown in Fig. \ref{fig:clustering}. In this pass, the verbs in the learning outcomes are identified using spaCy pos tagger\footnote{https://spacy.io/usage/v3} and WordNet synsets \cite{miller1995wordnet}, and then these verbs are used to determine the cluster to which the learning outcomes are most aligned with. While encountering verbs already available in the list, a straightforward approach is followed and the learning outcomes are assigned to the respective cluster. However, for learning outcomes with new verbs, an optimal cluster is determined based on the semantic similarity between the new verb and the verbs in the existing clusters. The best measure to calculate this similarity is determined as a result of the comparative analysis carried out between various knowledge-based and corpus-based semantic similarity measures. 
Three knowledge-based measures namely $wup$, $lch$, and $path$ are measured using WordNet ontology. In this ontology, there are more than one synsets for verbs hence it is necessary to identify the best synset. Given a verb pair, $wup$, $lch$, and $path$ select the first synset of the verbs, whereas $wup_max$, $lch_max$, $path_max$ identifies the synset that has the maximum similarity with the verb pairs. Gerz et al. \cite{gerz2016simverb} proposed SimVerb-3500 a benchmark dataset consisting of verb pairs with associated similarity values provided by annotators using crowd-sourcing techniques. The performance of six knowledge-based semantic similarity measures and word embeddings models (word2vec and GloVe) on the SimVerb-3500 is compared and the results are depicted in Fig. \ref{fig:simverb}. The pseudocode for this algorithm is provided in the Appendix. 
The best results are achieved by $wup\_max$ measure, hence used in the clustering process. 
\begin{figure*}
    \centering
    \includegraphics[width=0.8\textwidth]{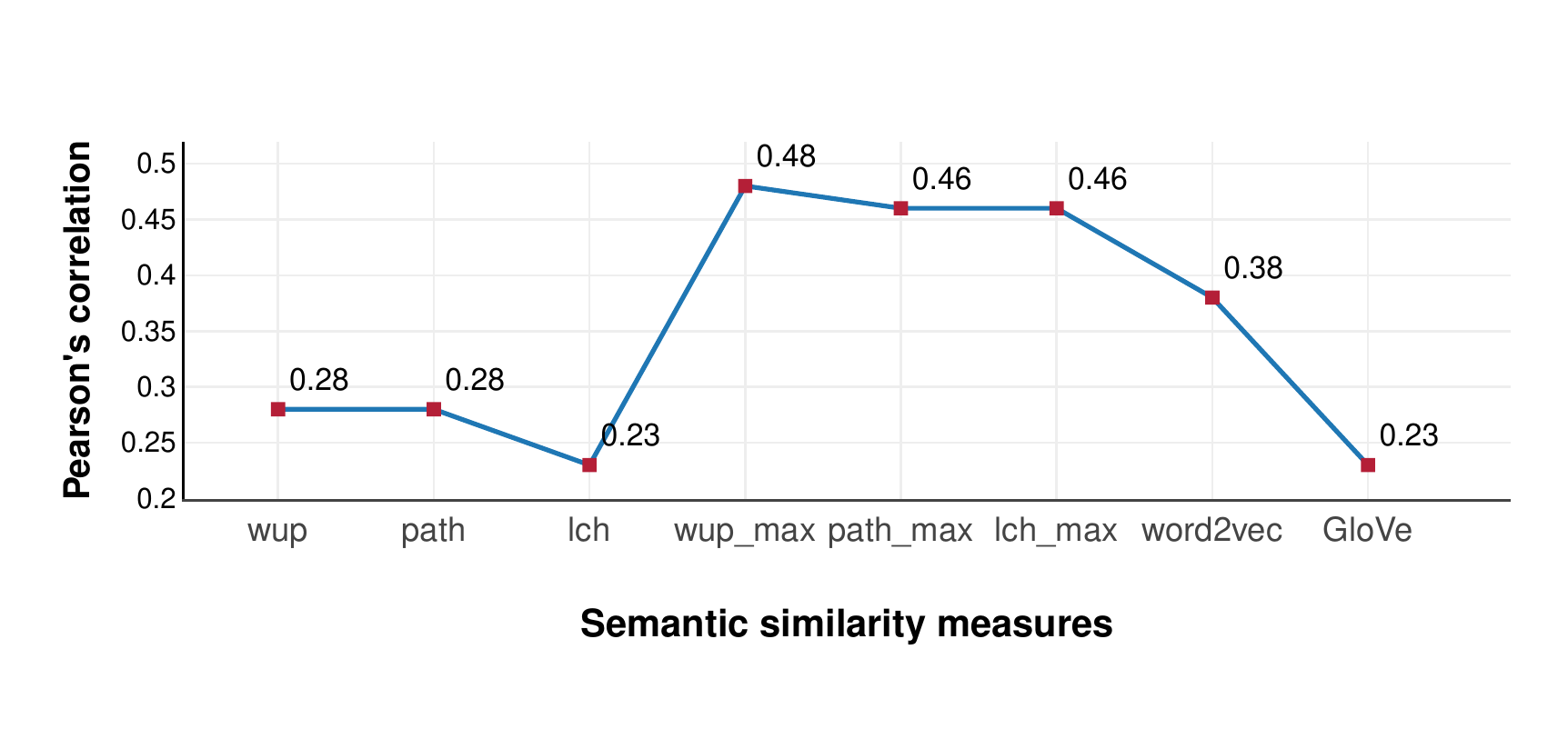}
    \caption{Pearson's correlation of various semantic similarity measures on Simverb3500 dataset. $wup$\cite{wu1994verbs}, $path$\cite{rada1989development}, $lch$\cite{leacock1998combining}, word2vec\cite{mikolov2013efficient}, GloVe\cite{pennington2014glove}}
    \label{fig:simverb}
\end{figure*}
Silhouette width is defined as "the measure of how much more similar a data point is to the points in its own cluster than to points in a neighboring cluster\cite{rousseeuw1987silhouettes}". The silhouette width of a verb $\mathcal{V}_i$ is measured as,

\begin{equation}
    \mathcal{S}_i = \frac{b_i-a_i}{max(a_i,b_i)}
\end{equation}
where, $a_i$ is the average distance between the verb and the verbs in its own cluster and $b_i$ is the average distance between the verb and the verbs in its neighboring cluster. The following equations formulate the calculation of these distance measures.\\ 
\noindent\begin{minipage}{.5\linewidth}
\begin{equation*}
  a_i = \frac{\displaystyle \sum_{\mathcal{V}_j \in C(\mathcal{V}_i)} d(\mathcal{V}_i,\mathcal{V}_j)}{|C(\mathcal{V}_j)|}
\end{equation*}
\end{minipage}%
\begin{minipage}{.5\linewidth}
\begin{equation*}
  b_i = \frac{\displaystyle \sum_{\mathcal{V}_k \in C_{neig}(\mathcal{V}_i)} d(\mathcal{V}_i,\mathcal{V}_k)}{|C_{neig}(\mathcal{V}_k)|}
\end{equation*}
\vspace{3mm}
\end{minipage}
$\mathcal{V}_j$ represent the verbs in the same cluster $C$ and $\mathcal{V}_k$ represent the verbs in the neighboring cluster $C_{neig}$, the function $d(\mathcal{V}_i,\mathcal{V}_j)$ calculates the distance between the verbs in this case the $wup\_measure$.
The value of $\mathcal{S}$ ranges from -1 to 1 where values closer to 1 indicate that the data point is assigned to the correct cluster. Based on this principle, when the model identifies a \textit{new\_verb} in a learning outcome, the silhouette width for each cluster is determined and the learning outcome is assigned to the cluster with maximum silhouette width. For learning outcomes with more than one verb, the verb assigned to the highest taxonomic level determines the final cluster value of the learning outcome in question. Since each cluster represents a corresponding level of competency in Bloom’s taxonomy, the final taxonomic similarity between the learning outcomes is measured as,
\begin{equation}\label{tax_eq}
    taxonomic\_similarity \hspace{2mm}(lo1,\hspace{1mm}lo2) = abs\hspace{2mm}( C_{lo1} - C_{lo2})
\end{equation},
where $C_{lo1}$ and $C_{lo2}$ represent the cluster ids of the learning outcomes in comparison calculated as,
\begin{equation}
    C_{loi} = max\hspace{2mm}(C(v_1), C(v_2).... C(v_n))
\end{equation}

where $n$ is the number of verbs in the learning outcome $loi$ and $C(v_n)$ represents the respective cluster of the verb $v_n$. For two courses having $m$ and $n$ number of learning outcomes, a $m \times n$ dimensional \textit{taxonomic\_similarity\_grid} is constructed and populated with the respective $taxonomic\_similarity$ values. 

\subsection{Pass 2: Semantic Similarity} 
Recent transformer-based language models generate contextual word embeddings that are fine-tuned for a specific NLP task, and various researchers have attempted to generate sentence embeddings by averaging the output from the final layer of the language models \cite{Zhang2020BERTScore:}. However, Reimers et al. \cite{reimers-2019-sentence-bert} established that these techniques yielded poor results in regression and clustering tasks like semantic similarity and proposed the Sentence BERT (SBERT) model to address this issue. SBERT is a modified version of BERT-based language models where a siamese network and triple network structures are added to the final layer of the BERT network to generate sentence embeddings that capture the semantic properties and thus can be compared using cosine similarity. Reimers et al. \cite{reimers-2019-sentence-bert} compared the performance of both SBERT and SRoBERTa language models in the STS \cite{shao2017hcti} and SICK \cite{marellisick} benchmark datasets. Owing to the fact that the model is specifically designed for semantic similarity tasks, the computational efficiency of the model architecture, and the significant performance achieved, SRoBERTa is used in this pass to measure the semantic similarity of the LOs. SRoBERTa uses the base RoBERTa-large model with 24 transformer blocks, 1024 hidden layers, 16 attention heads, and 340M trainable parameters with a final mean pooling layer. The model is trained on the AllNLI dataset which contains 1 million sentence pairs categorized into three classes namely ‘contradiction, entailment, and neutral’, and the training data of the STS benchmark dataset. The semantic similarity between the learning outcomes is determined by the cosine value between the embeddings as,
\begin{equation}\label{sem_eq}
    \displaystyle semantic\_similarity\hspace{2mm}( lo1,\hspace{1mm}lo2) = \frac{lo1 \centerdot lo2}{\sqrt{\displaystyle\sum_{i=1}^{n} lo1 } \sqrt{\displaystyle\sum_{i=1}^{n} lo2 }}
\end{equation}

The \textit{semantic\_similarity\_grid} with the dimension of $m \times n$ is formed and the $semantic\_similarity$ values are added to the cells. The output from the two initial passes are fed to the final pass for aggregation.

\begin{table*}
\caption{Survey results for the proposed benchmark dataset.}
\label{tab:survey}
\centering
\begin{tabular}{lllllll}
\toprule
\textbf{}  & \textbf{A1}                              & \textbf{A2}                                                      & \textbf{A3}                              & \textbf{A4}                              & \textbf{A5}                              & \textbf{\begin{tabular}[c]{@{}c@{}}Final\\ Annotation\end{tabular}}         \\ \midrule
\textbf{1} & {\color[HTML]{009901} \textbf{COMP4121}} & {\color[HTML]{009901} \textbf{COMP4121}}                         & {\color[HTML]{FE0000} \textbf{COMP4121}} & {\color[HTML]{009901} \textbf{COMP4121}} & {\color[HTML]{FE0000} \textbf{COMP4121}} & {\color[HTML]{009901} \textbf{COMP4121}} \\ 
\textbf{2} & {\color[HTML]{FE0000} \textbf{COMP5341}} & \cellcolor[HTML]{FFFFFF}{\color[HTML]{FE0000} \textbf{COMP5341}} & {\color[HTML]{FE0000} \textbf{COMP5341}} & {\color[HTML]{FE0000} \textbf{COMP5341}} & {\color[HTML]{FE0000} \textbf{COMP5341}} & {\color[HTML]{FE0000} \textbf{COMP5341}} \\ 
\textbf{3} & {\color[HTML]{FE0000} \textbf{COMP3321}} & \cellcolor[HTML]{FFFFFF}{\color[HTML]{009901} \textbf{COMP3321}} & {\color[HTML]{009901} \textbf{COMP3321}} & {\color[HTML]{FE0000} \textbf{COMP3321}} & {\color[HTML]{FE0000} \textbf{COMP3321}} & {\color[HTML]{FE0000} \textbf{COMP3321}} \\ 
\textbf{4} & {\color[HTML]{009901} \textbf{COMP4321}} & {\color[HTML]{009901} \textbf{COMP4321}}                         & {\color[HTML]{009901} \textbf{COMP4321}} & {\color[HTML]{FE0000} \textbf{COMP4321}} & {\color[HTML]{FE0000} \textbf{COMP4321}} & {\color[HTML]{009901} \textbf{COMP4321}} \\ 
\textbf{5} & {\color[HTML]{009901} \textbf{COMP3519}} & {\color[HTML]{009901} \textbf{COMP3519}}                         & {\color[HTML]{009901} \textbf{COMP3519}} & {\color[HTML]{009901} \textbf{COMP3519}} & {\color[HTML]{FE0000} \textbf{COMP3519}} & {\color[HTML]{009901} \textbf{COMP3519}} \\ 
\textbf{6} & {\color[HTML]{FE0000} \textbf{COMP5470}} & {\color[HTML]{009901} \textbf{COMP5470}}                         & {\color[HTML]{009901} \textbf{COMP5470}} & {\color[HTML]{FE0000} \textbf{COMP5470}} & {\color[HTML]{FE0000} \textbf{COMP5470}} & {\color[HTML]{FE0000} \textbf{COMP5470}} \\ 
\textbf{7} & {\color[HTML]{009901} \textbf{COMP5471}} & {\color[HTML]{009901} \textbf{COMP5471}}                         & {\color[HTML]{009901} \textbf{COMP5471}} & {\color[HTML]{FE0000} \textbf{COMP5471}} & {\color[HTML]{009901} \textbf{COMP5471}} & {\color[HTML]{009901} \textbf{COMP5471}} \\ 
\bottomrule
\end{tabular}
\end{table*}

\subsection{Pass 3: Aggregation}
The focus of the final pass is to provide flexibility in the aggregation process to enable the decision-making authorities to accommodate the variations in the administrative process across different institutions. Three important tunable parameters are provided to adjust the level of leniency offered by the decision-making authority in providing credits namely, `$impact$', `$sim\_threshold$', and `$lo\_threshold$'. Given the taxonomic similarity grid and the semantic similarity grid from Pass 1 and Pass 2 the $impact$ parameter determines the percentage of contribution of both the similarities to the overall similarity. The `$sim\_threshold$' defines the value above which two learning outcomes are considered to be similar, and finally, the `$lo\_threshold$' determines the number of learning outcomes that need to be similar in order to consider the courses in comparison to being similar. The higher the value of these three parameters the lesser the leniency in the decision-making process. 

The \textit{final\_similarity\_grid} is built by aggregating the values from the previous modules, in the ratio determined by the ‘$impact$’ parameter as shown in Fig. \ref{fig:archi}. The LOs along the rows of the grid belong to the receiving institution's course hence traversing along the rows, the maximum value in the cells is checked against the `$sim\_threshold$' value to determine if the LO in the row is similar to any LO in the columns. The course level similarity is derived by checking if the number of learning outcomes having a similar counterpart meets the `$lo\_threshold$'.

\begin{table}[]
\caption{Performance comparison of the proposed model with human annotation.}
\label{tab:results}
\centering
\begin{tabular}{cccc}
\toprule
\textbf{}          & \textbf{Human}                           & \textbf{\begin{tabular}[c]{@{}c@{}}Semantic\\ Similarity\end{tabular}} & \textbf{\begin{tabular}[c]{@{}c@{}}Proposed\\ Methodology\\ Neutral\end{tabular}} \\
\midrule
\textbf{1}         & {\color[HTML]{009901} \textbf{COMP4121}} & {\color[HTML]{009901} \textbf{COMP4121}}                               & {\color[HTML]{009901} \textbf{COMP4121}}                                          \\
\textbf{2}         & {\color[HTML]{FE0000} \textbf{COMP5341}} & \cellcolor[HTML]{FFFFFF}{\color[HTML]{FE0000} \textbf{COMP5341}}       & {\color[HTML]{FE0000} \textbf{COMP5341}}                                          \\
\textbf{3}         & {\color[HTML]{FE0000} \textbf{COMP3321}} & \cellcolor[HTML]{FFFFFF}{\color[HTML]{FE0000} \textbf{COMP3321}}       & {\color[HTML]{FE0000} \textbf{COMP3321}}                                          \\
\textbf{4}         & {\color[HTML]{009901} \textbf{COMP4321}} & {\color[HTML]{FE0000} \textbf{COMP4321}}                               & {\color[HTML]{009901} \textbf{COMP4321}}                                          \\
\textbf{5}         & {\color[HTML]{009901} \textbf{COMP3519}} & {\color[HTML]{FE0000} \textbf{COMP3519}}                               & {\color[HTML]{FE0000} \textbf{COMP3519}}                                          \\
\textbf{6}         & {\color[HTML]{FE0000} \textbf{COMP5470}} & {\color[HTML]{FE0000} \textbf{COMP5470}}                               & {\color[HTML]{FE0000} \textbf{COMP5470}}                                          \\
\textbf{7}         & {\color[HTML]{009901} \textbf{COMP5471}} & {\color[HTML]{009901} \textbf{COMP5471}}                               & {\color[HTML]{009901} \textbf{COMP5471}}                                          \\
\midrule
\textbf{\begin{tabular}[c]{@{}c@{}}Agreement\\ in \%\end{tabular}} & \textbf{}                                & 57.16                                                                & \textbf{85.74}    \\                       \bottomrule                                   
\end{tabular}
\end{table}

\section*{Dataset and Results}
\subsection{Benchmark Dataset}
One of the major challenges in the given field of research is the absence of benchmark datasets to compare the performance of the proposed system. Although there are existing pathways developed manually, previous research show that most of them are influenced by bias (based on the reputation of institutions, year of study, and so on) and administrative accommodations \cite{taylor2017multiple}. To create a benchmark dataset devoid of bias, a survey was conducted among domain experts to analyze and annotate the similarity between courses from two different institutions. A survey with learning outcomes of 7 pairs of courses (sending and receiving) from the computer science domain was distributed among instructors from the department of computer science at a comprehensive research university in Canada. In order to avoid bias, the names of both the institutions were anonymized and explicit instructions were provided to the annotators to assume a neutral position. Although the survey was circulated among 14 professors only 5 responses were received. This lack of responses is mainly attributed to the fact that most faculties are not involved in the transfer pathway development process and it is carried out widely as an administrative task. The survey questionnaire consisted of questions to mark the similarity between the courses over a scale of 1 to 10 and a binary response (\textit{`yes'} and \textit{`no'}) for whether or not credit should be offered to the receiving course. The course pairs were annotated with a final decision value based on the maximum number of \textit{`yes'} or \textit{`no'} responses received from the annotators. The results of the survey are tabulated in Table \ref{tab:survey} where the responses \textit{`yes'} and \textit{`no'} are color coded in \textit{`green'} and \textit{'red'} respectively. One of the interesting inferences from the survey results is the agreement between the responses in the threshold value of similarity above which the annotators were willing to offer credit. 4 out of 5 annotators offered credit only if they considered the similarity between the courses falls above 7 on the given scale of 1 to 10. It is also interesting to note that in spite of having no information other than the learning outcomes the annotators differed in their level of leniency which inspired the need to offer flexibility to control leniency in the proposed methodology. For example, from Table \ref{tab:survey} it is evident that while annotator \textit{`A2'} has followed a more lenient approach and offered credit for 6 out of 7 courses, annotator \textit{`A5'} has adopted a more strict approach by offering credit for only 1 out of the 7 courses.
\begin{table*}[ht!]
\caption{Performance of variations of proposed models with specific annotators.}
\label{tab:anno}
\small
\centering
\begin{tabular}{ccccc}
\toprule
\textbf{S.No}          & \textbf{\begin{tabular}[c]{@{}c@{}}Most lenient\\  annotator\end{tabular}} & \textbf{\begin{tabular}[c]{@{}c@{}}Proposed \\ methodology\\ Lenient\end{tabular}} &\textbf{\begin{tabular}[c]{@{}c@{}}Most strict\\  annotator\end{tabular}}& \textbf{\begin{tabular}[c]{@{}c@{}}Proposed \\ methodology\\ Strict\end{tabular}} \\
\midrule
\textbf{1}         & {\color[HTML]{009901} \textbf{COMP4121}}                                   & {\color[HTML]{009901} \textbf{COMP4121}}                                           & {\color[HTML]{FE0000} \textbf{COMP4121}}                                  & {\color[HTML]{009901} \textbf{COMP4121}}                                          \\
\textbf{2}         & {\color[HTML]{FE0000} \textbf{COMP5341}}                                   & {\color[HTML]{FE0000} \textbf{COMP5341}}                                           & {\color[HTML]{FE0000} \textbf{COMP5341}}                                  & {\color[HTML]{FE0000} \textbf{COMP5341}}                                          \\
\textbf{3}         & {\color[HTML]{009901} \textbf{COMP3321}}                                   & {\color[HTML]{009901} \textbf{COMP3321}}                                           & {\color[HTML]{FE0000} \textbf{COMP3321}}                                  & {\color[HTML]{FE0000} \textbf{COMP3321}}                                          \\
\textbf{4}         & {\color[HTML]{009901} \textbf{COMP4321}}                                   & {\color[HTML]{009901} \textbf{COMP4321}}                                           & {\color[HTML]{FE0000} \textbf{COMP4321}}                                  & {\color[HTML]{FE0000} \textbf{COMP4321}}                                          \\
\textbf{5}         & {\color[HTML]{009901} \textbf{COMP3519}}                                   & {\color[HTML]{009901} \textbf{COMP3519}}                                           & {\color[HTML]{FE0000} \textbf{COMP3519}}                                  & {\color[HTML]{FE0000} \textbf{COMP3519}}                                          \\
\textbf{6}         & {\color[HTML]{009901} \textbf{COMP5470}}                                   & {\color[HTML]{FE0000} \textbf{COMP5470}}                                           & {\color[HTML]{FE0000} \textbf{COMP5470}}                                  & {\color[HTML]{FE0000} \textbf{COMP5470}}                                          \\
\textbf{7}         & {\color[HTML]{009901} \textbf{COMP5471}}                                   & {\color[HTML]{009901} \textbf{COMP5471}}                                           & {\color[HTML]{009901} \textbf{COMP5471}}                                  & {\color[HTML]{009901} \textbf{COMP5471}}                                          \\
\midrule
\textbf{\begin{tabular}[l]{c@{}}Agreement\\ in \%\end{tabular}} & \textbf{}                                                                  & \textbf{85.74}                                                                       & \textbf{}                                                                 & \textbf{85.74}   \\
\bottomrule
\end{tabular}
\end{table*}
\subsection{Results} 
The results of the proposed model are provided in Table \ref{tab:results}. In order to provide context to the need for the proposed methodology, the results of the model are compared to the results obtained when only the semantic similarity of the learning outcomes is considered. The proposed model at a neutral setting achieves 85.74\% agreement with the human annotation obtained during the survey while the semantic similarity model achieves only 54.75\% agreement. For the neutral setting of the proposed model, the three parameters in the aggregation pass are set as follows. The $impact$ parameter is set at 30\% meaning that the semantic similarity contributes 70\% to the overall similarity and the taxonomic similarity contributes to the remaining 30\%. The $sim\_threshold$ is set at 65\%, meaning that the overall similarity should be more than 65\% in order for the learning outcomes to be similar to each other. The $lo\_threshold$ is set at 0.5 which considers that at least half of the available learning outcomes have similar counterparts. In order to demonstrate the options for flexibility, the proposed model is run by modifying the $lo\_threshold$ parameter. As shown in Table \ref{tab:anno} for the lenient setting the $lo\_threshold$ parameter is set at 0.33 and 0.66 and the model achieves an agreement of 85\% with the most lenient annotator and the most strict annotator, respectively. Similarly, lowering the $impact$ parameter makes the model more aligned to being strict and vice versa. However, it is important to understand that increasing the impact of taxonomic similarity to a higher percentage results in determining the overall similarity based on only two or three action words in the learning outcomes. Decreasing the $sim\_threshold$ will make the model more lenient and increasing it will make the model more strict. The results of the model at various parameter settings are provided in Table \ref{tab:S2} and Table \ref{tab:S3} in the Appendix.


\begin{figure*}
\centering
\includegraphics[width=\textwidth]{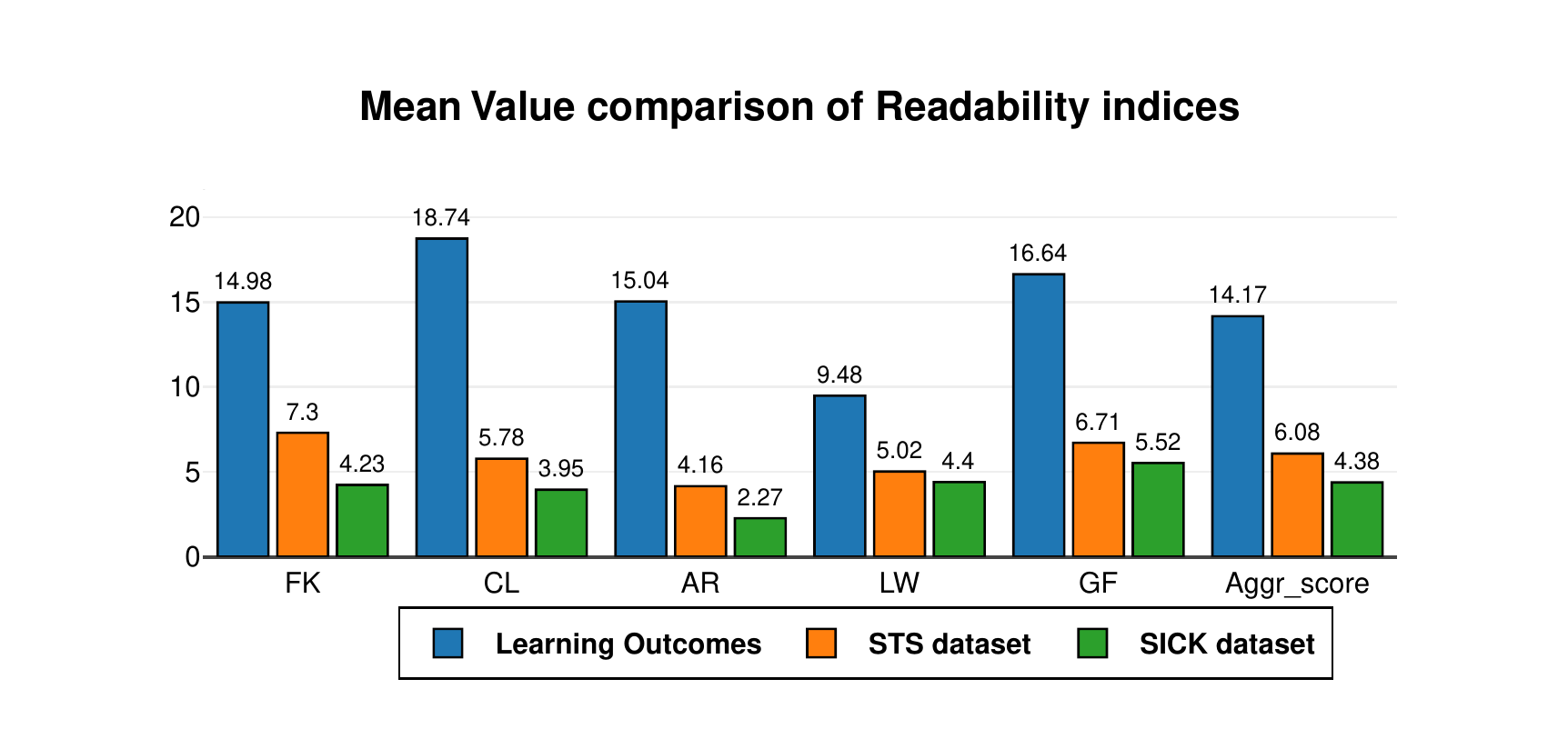}
\caption{Comparison of the readability indices of the learning outcomes used in the proposed dataset and the existing benchmark datasets (STS and SICK dataset).}
\label{fig:los}
\end{figure*}

\section*{Challenges and Future Works}
It is pertinent to understand that there are unique challenges to be addressed in the attempt to automate articulations. Although the existing transformer-based models achieve near-perfect results in benchmark datasets, a thorough understanding of these datasets brings to light one of their major shortcomings. Rogers et al. \cite{rogers2020primer} conclude in their survey with a clear statement on the limitations of the benchmark datasets, "As with any optimization method, if there is a shortcut in the data, we have no reason to expect BERT to not learn it. But harder datasets that cannot be resolved with shallow heuristics are unlikely to emerge if their development is not as valued as modeling work." Chandrasekaran et al. \cite{chandrasekaran2020domain} established the significant drop in performance of these models with the increase in complexity of sentences. The comparison of the complexity of the learning outcomes in the proposed benchmark dataset to the sentences in the STS dataset is shown in \ref{fig:los} , which clearly indicates that learning outcomes are more complex. Also, the introduction of domain-specific BERT models shows clear indications that though the transformer models are trained using significantly large corpora with millions of words in their vocabularies, a domain-specific corpus is required to achieve better results in domain-specific tasks. Learning outcomes are not only complex sentences but also contain domain-specific terminologies from various domains. Identifying these research gaps in semantic similarity methods is essential to contextualize and focus future research on addressing them. In addition to the technological challenges, it is also important to understand the challenges faced in the field of articulation. One of the approaches to enhance the performance of existing semantic similarity models is to train them with a large dataset of learning outcomes with annotated similarity values. However, learning outcomes are often considered to be intellectual properties of the instructors and are not publicly available. While almost all universities focus on building quality learning outcomes, most learning outcomes are either vague or don’t follow the structural requirements of learning outcomes \cite{schoepp2019state}. Even if a significant amount of learning outcomes are collected, annotation of their similarity requires expertise in subject matter and understanding of the articulation process. This annotation process is considerably more expensive than the annotation of English sentences as well. For example, one of the popular crowdsourcing platforms charges \$0.04 for annotators with no specification and \$0.65 for an annotator with at least a Master’s degree. Furthermore, the selection of annotators with the required expertise needs manual scrutiny using preliminary questionnaires and surveys which makes the process time-consuming. Finally, articulation agreements are developed across different departments and it is imminent to provide a clear understanding of the model and its limitations to encourage automation of the process. While the proposed model accommodates most of these challenges by allowing transparency and flexibility in the assessment of credit transfer, there is scope for improvement, and the future research works will focus on developing domain-specific corpora, tuning the semantic similarity models with the aid of datasets and also on ways to improve the generation of learning outcomes through automation.
\section*{Conclusion}
The assessment of transfer credit in the process of student mobility is considered to be one of the most crucial and time-consuming tasks, and across the globe, various steps have been taken to mitigate this process. With significant research and advancements in the field of natural language processing, this research article attempts to automate the articulation process by measuring the semantic and taxonomic similarity between learning outcomes. The proposed model uses a recent transformer-based language model to measure the semantic similarity and a clustering-inspired methodology is proposed to measure the taxonomic similarity of the learning outcomes. The model also comprises a flexible aggregation module to aggregate the similarity between learning outcomes to course-level learning outcomes. A benchmark dataset is built by conducting a survey among academicians to annotate the similarity and transfer credit decisions between courses from two different institutions. The results of the proposed model are compared with those of the benchmark dataset at different settings of leniency. The article also identifies the technical and domain-specific challenges that should be addressed in the field of automating articulation.
\subsection*{Data Availability} The source code used to generate all of
the data associated with this article is provided in this 
GitHub repository (https://github.com/Dhivya-C/Transfer-credit-assessment).
\subsection*{ACKNOWLEDGMENT} The authors would like to extend our gratitude to the research team in the DaTALab at Lakehead University for their support and Arunim Garg for the valuable feedback and revisions on this publication. We would also like to thank Lakehead University, CASES and the Ontario Council for Articulation and Transfer, without their support this research would not have been possible.

\bibliography{pnas-sample}

\appendix
\onecolumn
   \begin{algorithm}
        \caption{Best semantic similarity measure for verbs}\label{alg:bestmeasure}
        \begin{flushleft}
        \textbf{Input:} $\mathcal{D}_{verb} [v_1,v_2,s]$ = SimVerb3500 dataset containing verb pairs ($v_1,v_2$) and associated semantic similarity benchmark values $s$, \\
        $wv$ = pretrained $word2vec$ word embedding model\\
        $gl$ = pretrained $GloVe$ word embedding model\\
        $wn$ = $WordNet$ lexical knowledge base\\
        \textbf{Output:} $best\_measure$ = the semantic similarity measure that achieves the best correlation to the benchmark values.
        \end{flushleft}
        
        \begin{algorithmic}[1]
        \State $\mathcal{S} \gets \varnothing$
        \Procedure{verb\_similarity}{}
            \ForEach{$(v_1,v_2,s) \in \mathcal{D}_{verb}$ }
                \State $\mathcal{S} +\!= s$
                \State $w1\gets wv.embedding(v_1)$
                \State $w2\gets wv.embedding(v_2)$
                \State $g1\gets gl.embedding(v_1)$
                \State $g2\gets gl.embedding(v_2)$
                \State $synsets1[syn_{11}... syn_{1i}]\gets wn.get\_synsets(v_1,POS:verb)$ \Comment{List of synsets}
                \State $synsets2[syn_{21}....syn_{2j}]\gets wn.get\_synsets(v_2,POS:verb)$ \Comment{List of synsets}
                \ForEach{$syn_{1i} \in synsets1 $}
                    \ForEach{$syn_{2j} \in synsets2 $}
                        \State $wup\_sim\_list[syn_{1i}][syn_{2j}] \gets cal\_wup(syn_1,syn_2)$ \Comment{wup\_similarity}
                        \State $path\_sim\_list[syn_{1i}][syn_{2j}]  \gets cal\_path(syn_1,syn_2)$ \Comment{path\_similarity}
                        \State $lch\_sim\_list[syn_{1i}][syn_{2j}]  \gets cal\_lch(syn_1,syn_2)$ \Comment{lch\_similarity}
                    \EndFor
                \EndFor    
                \State $wup\_sim\gets max(wup\_sim\_list[n])$ 
                \State $path\_sim\gets max(path\_sim\_list[n])$
                \State $lch\_sim\gets max(lch\_sim\_list[n])$
                \State $word2vec\_sim \gets wv.similarity(w1,w2)$
                \State $glove\_sim \gets gl.similarity(g1,g2)$
            \EndFor
            \State $measure\_list +\!= wup\_sim, path\_sim, lch\_sim,word2vec\_sim, glove\_sim$           
            \ForEach{$m \in measure\_list$}
                \State $cor \gets get\_pearson\_correlation(S,m)$
            \EndFor
            \State $best\_measure \gets max(cor)$
            \State \textbf{return} $best\_measure$
        \EndProcedure
    \end{algorithmic}
    \end{algorithm}
    \begin{algorithm}
        \caption{Pass\_1: Taxonomic Similarity}\label{alg:pass1}
        \begin{flushleft}
        \textbf{Input:} $\mathcal{S}{lo\_list}$ = List of $m$ learning outcomes from the sending institution.\\
        $\mathcal{R}{lo\_list}$ = List of $n$ learning outcomes from the receiving institution.\\
        $\mathcal{C}luster\_list = \{C_1,C_2,C_3,C_4,C_5,C_6\}$, List of six clusters initialized with illustrative verbs, such that $C_1 = \{v_1,v_2...v_n\}$ is a list of n verbs  \\
        \textbf{Output:} $taxonomic\_similarity\_grid$ = A $m\times n$ grid containing taxonomic similarity values between sending and receiving learning outcomes
        \end{flushleft}
        \begin{algorithmic}[1]
        \Procedure{build\_taxonomic\_similarity\_grid}{}
            \ForEach{$(slo) \in \mathcal{S}{lo\_list}$}
                \State $slo\_verbs \gets detect\_verbs(slo)$ \Comment{Find the verbs in the learning outcomes}
                \ForEach{$sv \in slo$}
                    \If{$sv \in \mathcal{C}luster\_list$}
                        \State $\mathcal{B}(sv) \gets get\_index(C_i)$ \Comment{If the verb is among the predefined verbs \\\hspace{9cm} assign the corresponding cluster id.}
                    \Else
                        \State $\mathcal{B}(sv) \gets get\_BestCluster(sv)$ \Comment{Assign the cluster id of the best cluster\\ \hspace{9cm} based on the silhouette width}
                    \EndIf
                \EndFor
            \State $\mathcal{B}(slo) \gets max\big(\mathcal{B}(sv_1),....\mathcal{B}(sv_n)\big)$
            \EndFor
            \\\Comment{Repeat for learning outcomes from \\ \hspace{9cm} receiving institution}
            \ForEach{$(rlo) \in \mathcal{R}{lo\_list}$}
                \State $rlo\_verbs \gets detect\_verbs(rlo)$ 
                \ForEach{$rv \in rlo$}
                    \If{$rv \in \mathcal{C}luster\_list$}
                        \State $\mathcal{B}(rv) \gets get\_index(C_i)$ 
                    \Else
                        \State $\mathcal{B}(rv) \gets get\_BestCluster(rv)$ 
                    \EndIf
                \EndFor
            \State $\mathcal{B}(rlo) \gets max(\mathcal{B}(rv_1),....\mathcal{B}(rv_n)$
            \EndFor
            \\\Comment{Calculate taxonomic similarity}
            \ForEach{$(slo) \in \mathcal{S}{lo\_list}$}
                \ForEach{$(rlo) \in \mathcal{R}{lo\_list}$}
                    \State $taxonomic\_similarity\_grid [slo][rlo] \gets abs\big(\mathcal{B}(rlo) - \mathcal{B}(slo)\big)$
                \EndFor
            \EndFor
            \State \textbf{return} $taxonomic\_similarity\_grid$
        \EndProcedure
        \end{algorithmic}
    \end{algorithm}

\begin{algorithm}
        \caption{Pass\_2: Semantic Similarity}\label{alg:pass2}
        \begin{flushleft}
        \textbf{Input:} $\mathcal{S}{lo\_list}$ = List of $m$ learning outcomes from the sending institution.\\
        $\mathcal{R}{lo\_list}$ = List of $n$ learning outcomes from the receiving institution.\\
        $model =$ Pretrained and Fine tuned SRoBERTa transformer-based semantic similarity model
        \textbf{Output:} $semantic\_similarity\_grid$ = A $m\times n$ grid containing semantic similarity values between sending and receiving learning outcomes
        \end{flushleft}
        \begin{algorithmic}[1]
        \Procedure{build\_semantic\_similarity\_grid}{}
            \ForEach{$slo,rlo \in \mathcal{S}{lo\_list}\mathcal{R}{lo\_list}$}
               \State $\vec{slo} \gets model.sentence\_vectors(slo)$ \Comment{Get vector representation of \\\hspace{9cm}learning outcomes}
               \State $\vec{rlo} \gets model.sentence\_vectors(rlo)$\\
               \Comment{Calculate semantic similarity}
               \State $semantic\_similarity\_grid [slo][rlo] \gets cos(\vec{slo},\vec{rlo})$ 
            \EndFor
            \State \textbf{return} $semantic\_similarity\_grid$
        \EndProcedure
    \end{algorithmic}
    \end{algorithm}

    \begin{algorithm}
        \caption{Pass\_3: Aggregation}\label{alg:pass3}
        \begin{flushleft}
        \textbf{Input:} $SSG$ = $\mathit{semantic\_similarity\_grid}$, a $m\times n$ grid containing semantic similarity values between sending and receiving learning outcomes\\
         $TSG$ = $taxonomic\_similarity\_grid$, a $m\times n$ grid containing taxonomic similarity values between sending and receiving learning outcomes\\
        $\alpha =$ parameter which determines the ratio of two similarity value in the overall similarity\\
        $\beta =$ value above which two learning outcomes are considered similar\\
        $\gamma =$ number of learning outcomes with similar counterparts\\
        \textbf{Output:} $\mathcal{TC} =$ Final Credit Decision
        \end{flushleft}
        \begin{algorithmic}[1]
        \Procedure{course\_level\_similarity}{}
            \ForEach{$row, col \in SSG $}
                \ForEach{$row, col \in TSG $}
                    \State $final\_sim \gets \!
                    \begin{aligned}[t]
                    \big(\alpha \times SSG [row,col] +\\  (1-\alpha) \times TSG [row,col]\big) 
                    \end{aligned}$
                    \If{$final\_sim >= \beta$}
                        \State $final\_similarity\_grid[row,column] \gets TRUE$
                    \Else
                        \State $final\_similarity\_grid[row,column] \gets FALSE$
                    \EndIf
                \EndFor
            \EndFor
            \ForEach{$row \in final\_similarity\_grid$}
                \If {$Count(TRUE) >= \gamma$}
                   \State $\mathcal{TC} = Yes$
                \Else 
                    \State $\mathcal{TC} = No$
                \EndIf
            \EndFor
            \State \textbf{return} $\mathcal{TC}$
        \EndProcedure
        \end{algorithmic}
    \end{algorithm}

\begin{table}[]
    \centering
    \caption{Formula of the semantic similarity distance measures taken into consideration for determining verb similarity}
    \begin{tabular}{ccc}
        \toprule
        \textbf{S.No} &\textbf{Semantic Similarity Measure}          & \textbf{Formula}\\
        \midrule
         1   & $ sim_{path}(t_1,t_2) $ &  $\displaystyle \frac{1}{1+min\_len(t_1,t_2)} $   \\ 
          \\
        2   & $sim_{wup}(t_1,t_2)$        & 
            $ {\displaystyle \frac{2depth(t_{lcs})}{depth(t_1)+depth(t_2)}}$      \\ \\
        3   & $sim_{lch}(t_1,t_2)$ & $ {\displaystyle -\log} \frac{min\_len(t_1,t_2)}{2 \times max\_depth}$ \\\\
        \bottomrule
    \end{tabular}
    \label{tab:S1}
\end{table}

\begin{table}[]
\centering
\caption{Performance of the proposed model on varying $sim\_threshold$ parameter}
\label{tab:S2}
\begin{tabular}{ccccc}
\toprule
\textbf{S.No.}                                                     & \textbf{\begin{tabular}[c]{@{}c@{}}Human\\ Annotation\end{tabular}} & \textbf{\begin{tabular}[c]{@{}c@{}}sim\_threshold\\ = 60\end{tabular}} & \textbf{\begin{tabular}[c]{@{}c@{}}sim\_threshold\\ = 65\end{tabular}} & \textbf{\begin{tabular}[c]{@{}c@{}}sim\_threshold\\ = 70\end{tabular}} \\
\midrule
\textbf{1}                                                         & {\color[HTML]{009901} \textbf{COMP4121}}                            & {\color[HTML]{009901} \textbf{COMP4121}}                               & {\color[HTML]{009901} \textbf{COMP4121}}                               & {\color[HTML]{009901} \textbf{COMP4121}}                               \\
\textbf{2}                                                         & {\color[HTML]{FE0000} \textbf{COMP5341}}                            & {\color[HTML]{009901} \textbf{COMP5341}}                               & {\color[HTML]{FE0000} \textbf{COMP5341}}                               & {\color[HTML]{FE0000} \textbf{COMP5341}}                               \\
\textbf{3}                                                         & {\color[HTML]{FE0000} \textbf{COMP3321}}                            & {\color[HTML]{009901} \textbf{COMP3321}}                               & {\color[HTML]{009901} \textbf{COMP3321}}                               & {\color[HTML]{009901} \textbf{COMP3321}}                               \\
\textbf{4}                                                         & {\color[HTML]{009901} \textbf{COMP4321}}                            & {\color[HTML]{009901} \textbf{COMP4321}}                               & {\color[HTML]{009901} \textbf{COMP4321}}                               & {\color[HTML]{FE0000} \textbf{COMP4321}}                               \\
\textbf{5}                                                         & {\color[HTML]{009901} \textbf{COMP3519}}                            & {\color[HTML]{009901} \textbf{COMP3519}}                               & {\color[HTML]{009901} \textbf{COMP3519}}                               & {\color[HTML]{FE0000} \textbf{COMP3519}}                               \\
\textbf{6}                                                         & {\color[HTML]{FE0000} \textbf{COMP5470}}                            & {\color[HTML]{009901} \textbf{COMP5470}}                               & {\color[HTML]{FE0000} \textbf{COMP5470}}                               & {\color[HTML]{FE0000} \textbf{COMP5470}}                               \\
\textbf{7}                                                         & {\color[HTML]{009901} \textbf{COMP5471}}                            & {\color[HTML]{009901} \textbf{COMP5471}}                               & {\color[HTML]{009901} \textbf{COMP5471}}                               & {\color[HTML]{009901} \textbf{COMP5471}}                               \\
\midrule
\textbf{\begin{tabular}[c]{@{}c@{}}Agreement\\ in \%\end{tabular}} & \textbf{}                                                           & 57.14                                                                  & \textbf{85.71}                                                         & 57.14        \\
\bottomrule
\end{tabular}
\end{table}
\begin{table}[]
\centering
\caption{Performance of the proposed model on varying impact parameter}
\label{tab:S3}
\begin{tabular}{ccccc}
\toprule
\textbf{S.No.}                                                     & \textbf{\begin{tabular}[c]{@{}c@{}}Human\\ Annotation\end{tabular}} & \textbf{\begin{tabular}[c]{@{}c@{}}impact\\ = 20\end{tabular}} & \textbf{\begin{tabular}[c]{@{}c@{}}impact\\ = 30\end{tabular}} & \textbf{\begin{tabular}[c]{@{}c@{}}impact\\ = 40\end{tabular}} \\
\midrule
\textbf{1}                                                         & {\color[HTML]{009901} \textbf{COMP4121}}                            & {\color[HTML]{009901} \textbf{COMP4121}}                       & {\color[HTML]{009901} \textbf{COMP4121}}                       & {\color[HTML]{009901} \textbf{COMP4121}}                       \\
\textbf{2}                                                         & {\color[HTML]{FE0000} \textbf{COMP5341}}                            & {\color[HTML]{FE0000} \textbf{COMP5341}}                       & {\color[HTML]{FE0000} \textbf{COMP5341}}                       & {\color[HTML]{009901} \textbf{COMP5341}}                       \\
\textbf{3}                                                         & {\color[HTML]{FE0000} \textbf{COMP3321}}                            & {\color[HTML]{009901} \textbf{COMP3321}}                       & {\color[HTML]{009901} \textbf{COMP3321}}                       & {\color[HTML]{009901} \textbf{COMP3321}}                       \\
\textbf{4}                                                         & {\color[HTML]{009901} \textbf{COMP4321}}                            & {\color[HTML]{FE0000} \textbf{COMP4321}}                       & {\color[HTML]{009901} \textbf{COMP4321}}                       & {\color[HTML]{009901} \textbf{COMP4321}}                       \\
\textbf{5}                                                         & {\color[HTML]{009901} \textbf{COMP3519}}                            & {\color[HTML]{009901} \textbf{COMP3519}}                       & {\color[HTML]{009901} \textbf{COMP3519}}                       & {\color[HTML]{009901} \textbf{COMP3519}}                       \\
\textbf{6}                                                         & {\color[HTML]{FE0000} \textbf{COMP5470}}                            & {\color[HTML]{FE0000} \textbf{COMP5470}}                       & {\color[HTML]{FE0000} \textbf{COMP5470}}                       & {\color[HTML]{009901} \textbf{COMP5470}}                       \\
\textbf{7}                                                         & {\color[HTML]{009901} \textbf{COMP5471}}                            & {\color[HTML]{009901} \textbf{COMP5471}}                       & {\color[HTML]{009901} \textbf{COMP5471}}                       & {\color[HTML]{009901} \textbf{COMP5471}}                       \\
\midrule
\textbf{\begin{tabular}[c]{@{}c@{}}Agreement\\ in \%\end{tabular}} & \textbf{}                                                           & 71.42                                                          & \textbf{85.71}                                                 & 57.14             \\
\bottomrule
\end{tabular}
\end{table}
\end{document}